# Photoelectric Factor Prediction Using Automated Learning and Uncertainty Quantification

Khalid L. Alsamadony[1*], Ahmed Ibrahim[1,2**], Salaheldin Elkatatny[1,2], Abdulazeez Abdulraheem[1,2]

[1]Petroleum Engineering Department, King Fahd University of Petroleum & Minerals, Saudi Arabia.
[2]Center for Integrative Petroleum Research, King Fahd University of Petroleum & Minerals, Saudi Arabia.
*E-mail: khalidalsamadony@outlook.com    **E-mail: ahmed.ibrahim@kfupm.edu.sa

## Abstract

The photoelectric factor (PEF) is an important well logging tool to distinguish between different types of reservoir rocks because PEF measurement is sensitive to elements with high atomic number. Furthermore, the ratio of rock minerals could be determined by combining PEF log with other well logs.

However, PEF log could be missing in some cases such as in old well logs and wells drilled with barite-based mud. Therefore, developing models for estimating missing PEF log is essential in those circumstances. In this work, we developed various machine learning models to predict PEF values using the following well logs as inputs: bulk density (RHOB), neutron porosity (NPHI), gamma ray (GR), compressional and shear velocity.

The predictions of PEF values using adaptive-network-fuzzy inference system (ANFIS) and artificial neural network (ANN) models have errors of about 16% and 14% average absolute percentage error (AAPE) in the testing dataset, respectively. Thus, a different approach was proposed that is based on the concept of automated machine learning. It works by automatically searching for the optimal model type and optimizes its hyperparameters for the dataset under investigation.

This approach selected a Gaussian process regression (GPR) model for accurate estimation of PEF values. The developed GPR model decreases the AAPE of the predicted PEF values in the testing dataset to about 10% AAPE. This error could be further decreased to about 2% by modeling the potential noise in the measurements using the GPR model.

## 1 Introduction

The photoelectric factor (PEF) log is a robust tool for identifying lithology. This is because PEF measurement is mainly affected by rocks with high atomic number, while being insensitive to variations in porosity, fluid type and saturation. Therefore, PEF log capability to identify the lithology is not diminished in gas-bearing zones, where it may be challenging to distinguish between lithologies using only formation density and neutron logs. One more advantage of PEF log is that it can distinguish between clean (i.e., low shale content) sandstone and clean limestone, while gamma ray log cannot clearly demystify the difference between them. PEF ($P_e$) is proportional to the atomic number as shown in the next empirical relationship (Eq.1), where $z$ is the atomic number (Atlas, 1982; Bassiouni & others, 1994; Ellis & Singer, 2007).

$$P_e = \left(\frac{z}{10}\right)^{3.6} \qquad (1)$$

PEF log may be missing in some cases because of instrument failure, old logs missing this tool, and economic constraints. Moreover, PEF log could be mismeasured and misinterpreted in wells drilled with barite and in wells that have heavy elements even in small quantity. Remeasuring PEF log is often impractical because of well casing or limited budget. Therefore, previous studies utilized machine learning techniques to predict PEF log due to its importance in improving formation evaluation (Akinnikawe et al., 2018; Asoodeh & Shadizadeh, 2015).

**1.1 PEF Logs as Inputs for Machine Learning Models.**
PEF log is an important input parameter for many machine learning models to classify lithology and predict other valuable parameters such as density, porosity, water saturation, sonic and geochemical logs. PEF was used for training classifiers (k-nearest neighbors (KNN), support vector machine (SVM), random forest (RF), decision tree (DT), XGBoost, LightGBM, artificial neural network (ANN) and rough set theory rules) to predict lithofacies and electrofacies (Amir et al., 2020; Hossain et al., 2020; Merembayev et al., 2021). Moreover, it is an input for several models (ANN, SVM, Super Learner, XGBoost, CatBoost, LightGBM and AdaBoost) to estimate water saturation (Hadavimoghaddam et al., 2021; Miah et al., 2020).

PEF log was utilized to generate sonic logs (compressional and shear velocity) using ANN, recurrent neural networks (RNNs), RF, ensemble method (combination of several models) and LightGBM, where data preprocessing and feature engineering had a significant impact on the accuracy (Tatsipie & Sheng, 2021; Yu et al., 2021). Another study proposed fast fuzzy modeling method (FFMM) to predict sonic and density logs using PEF log as an input, where the proposed method had similar accuracy to ANN and fuzzy logic (FL), but with less computational and storage cost (Bahrpeyma et al., 2013).

PEF log is an important input in predicting geochemical logs (Al, Ca, Fe, Na, and Si) using the following methods: SVM, ANN, RF, AdaBoost and XGBoost (Blanes de Oliveira & de Carvalho Carneiro, 2021). Moreover, it is one of the inputs for long short-term memory (LSTM) model to output neutron porosity and resistivity logs (Tatsipie & Sheng, 2021).

**1.2 Prediction of PEF Logs**
Tatsipie and Sheng trained LSTM model to predict PEF log using the following inputs: gamma ray, bulk density, depth and well coordinates. PEF log was predicted with high errors relative to other predicted logs in this study, which was justified that the trained LTSM model could not capture the complex relation between PEF, gamma ray and bulk density (Tatsipie & Sheng, 2021).

Rostamian et al., predicted PEF, bulk density and neutron porosity logs using several machine learning methods: multilinear regression, deep neural network (DNN), DT, KNN, RF, and XGBoost. These models receive corrected gamma ray (CGR), caliper (CALI), deep latero log (LLD), shallow latero log (LLS), micro-spherical focused log (MSFL) and sonic transit time, where all the well logs were normalized (Z-score transformation). XGBoost, RF and KNN generated the best results, in addition their hyperparameters and selected inputs were optimized by genetic algorithm. This optimization slightly improved the accuracy and allowed reducing the required number of inputs, while predicting the missing logs with acceptable accuracy (Rostamian et al., 2022).

**1.3 Novelty**
The objective of this study is to develop machine learning models for estimating PEF values for a heterogenous dataset that was collected from a field in the middle east. In our dataset, the PEF values range from (2.1) to (13.4), while they range from about (1) to (7.8), (2.5) to (5.5) and (1) to (4.9) in the datasets of the previous studies (Rostamian et al., 2022; Tatsipie & Sheng, 2021; Wu et al., 2021).

The previous studies have predicted PEF values with relatively high error. Rostamian et al., predicted PEF values that have about 10 times more error than the prediction of neutron porosity (NPHI) according to the AAPE (Rostamian et al., 2022). Tatsipie and Sheng concluded that the predicted PEF log has the worst accuracy compared to the other predicted well logs (Tatsipie & Sheng, 2021). Our results, despite being based on a different dataset, confirm that predicting PEF log accurately could be challenging, hence this seems related to the prediction of PEF logs regardless of the dataset. For this reason, a method based on automated machine learning is proposed for better accuracy and efficient implementation of machine learning tools through removing the time-consuming manual steps. Additionally, we utilize uncertainty quantification to enhance the reliability of machine learning methods by modeling the potential noise that is usually in the measurements of actual cases at varying levels.

## 2 Methodology

In this work, we developed three machine learning models to predict PEF log, which are adaptive neuro-fuzzy inference system (ANFIS), artificial neural network (ANN) and Gaussian process regression (GPR). The inputs of these models are 5 logs: bulk density (RHOB), neutron porosity (NPHI), gamma ray (GR), compressional and shear velocity. The accuracy of the trained models is evaluated using the common metrics: average (mean) absolute percentage error (AAPE) (Eq.2) and Pearson correlation coefficient (R) (Eq.3) between the predicted and the actual PEF values. For the uncertainty quantification (UQ) model, AAPE refers to the percentage of the predicted PEF values that are outside the confidence interval (CI).

$$\text{AAPE} = \frac{100}{n} \times \sum_{i=1}^{n} \left| \frac{actual\ PEF_i - predicted\ PEF_i}{actual\ PEF_i} \right| \qquad (2)$$

$$R = \frac{\sum_{i=1}^{n}(x_i - \bar{x})(y_i - \bar{y})}{\sqrt{\sum_{i=1}^{n}(x_i - \bar{x})^2}\sqrt{\sum_{i=1}^{n}(y_i - \bar{y})^2}} \qquad (3)$$

where $n$ is the number of datapoints, $x_i$ is the actual PEF value, $\bar{x}$ is the average of the actual PEF values, $y_i$ is the predicted PEF value, and $\bar{y}$ is the average of the predicted PEF values.

### 2.1 Data Description

The dataset consists of 9399 datapoints from a field in the middle east such that 4922 datapoints were used for training the models, 895 datapoints for the initial testing, and 2603 datapoints for the final validation of the models.

Statistical analysis of the dataset is presented in Table 1, where the heterogeneity of the dataset can be observed. Specifically, the minimum and the maximum values of GR (3, 381) and NPHI (0, 0.44) indicate that the dataset contains different types of zones with distinct properties. The kurtosis values of GR (9.75) and NPHI (4.67) show that they have many extreme values. R-values depict that RHOB is the most linearly correlated to PEF (-0.37), that is followed by compressional velocity (0.18). R value measures the linear correlation between two variables, where value of one indicates a perfect proportional linear correlation, value of negative one indicates a perfect inversely proportional linear correlation, and zero indicates that there is no linear correlation between the two variables.

*Table 1* Statistical description of the dataset

| Statistical Analysis | RHOB | NPHI | GR | Compressional | Shear | PEF |
|---|---|---|---|---|---|---|
| Min | 2.18 | -0.01 | 3.17 | 40.31 | 60.82 | 2.14 |
| Max | 3.00 | 0.44 | 381.90 | 83.04 | 147.35 | 13.39 |
| Mean | 2.71 | 0.08 | 55.80 | 53.34 | 93.50 | 5.77 |
| Mode | 2.85 | 0.02 | 19.97 | 43.69 | 82.84 | 4.21 |
| Median | 2.59 | 0.22 | 192.53 | 61.67 | 104.08 | 7.76 |
| Standard deviation | 0.17 | 0.06 | 49.14 | 8.86 | 11.68 | 2.05 |
| Coef. of variation | 0.06 | 0.80 | 0.88 | 0.17 | 0.12 | 0.36 |
| Skewness | -0.43 | 1.21 | 2.29 | 0.75 | 1.23 | 0.83 |
| Kurtosis | 1.98 | 4.67 | 9.75 | 2.49 | 4.46 | 3.21 |
| R (PEF) | -0.37 | 0.10 | 0.03 | 0.18 | 0.13 | 1.00 |

Before the training process, the inputs and the PEF values were normalized to be between zero and one using their minimum and maximum values (Eq.4). However, the predicted PEF values were restored to the original scale (denormalized), so we can evaluate the accuracy of the models using AAPE and R. The dataset was checked for erroneous values such as zeros, negatives, and very extreme unreasonable values. Moreover, we attempted to remove the outliers using common statistical rules such as 1.5 IQR and 3 sigma (standard deviation). However, the accuracy of ANN model did not improve, but it slightly decreased. This probably due to the heterogenous nature of our dataset, where these extreme values represent some layers. Thus, removing them hinders the predictive capability of the artificial intelligence models. Additionally, numerous feature engineering and data transformation methods were applied, but they were not effective at reducing the error, as they only reduced AAPE by about 1% in the best case. As a consequence, a different approach was implemented utilizing automated learning and uncertainty quantification.

$$Normalized\ (x) = \frac{x - min(x)}{max(x) - min(x)} \tag{4}$$

**2.2 Models Description**

In this section, we briefly introduce the three machine learning methods that have been used to generate the results in our work, which are ANFIS, ANN and GPR. Moreover, we cite some references that explain the theory of these methods in detail and their different applications.

Adaptive-network-based fuzzy inference system (ANFIS) can be used as a supervised learning method since it is capable of modeling non-linear functions. Its prediction (inference) is based on Fuzzy rules (if-then rules) to map inputs of the system to outputs, thus it can resemble human knowledge and reasoning. These rules (if-then) are implemented using membership functions. This is the main difference between ANFIS and ANNs (Jang, 1993). Several studies used ANFIS to predict rock properties and well logs (Desouky et al., 2021; Gamal et al., 2021; Gowida et al., 2021).

ANNs have been widely used for classification and regression tasks. The basic component of ANN is the neuron, which forms the hidden layers. The hidden layers map the input layer to the output layer, and each hidden layer is followed by a transfer function to learn non-linear features. The neurons are associated with weights that are updated during the training process according to the backpropagated errors and the training algorithm e.g., gradient descent (Gurney, 1997). Several studies used ANN to predict rock properties and well logs (Asante-Okyere et al., 2018; Desouky et al., 2021; Gamal et al., 2021; Gowida et al., 2021; Hiba et al., 2022).

The general concept of GPR assumes that datapoints with similar input values are expected to have similar output values. This concept is applied using covariance (kernel) functions that model these relations of similarity (Rasmussen & Williams, 2006). Due to this, kriging that is the well-known method used in geo-statistics to interpolate the missing values based on the established spatial covariance or variogram, is considered a GPR method. GPR is less commonly used than ANN to predict rock properties and well logs (Asante-Okyere et al., 2018; Onalo et al., 2020).

## 3 Results

**3.1 ANFIS Results**

To develop the ANFIS model, we used subtractive clustering (Sugeno fuzzy) algorithm that searches for data clusters, from which membership functions and rules are constructed (Chiu, 1994; Yager & Filev, 1994). The main hyperparameter of the subtractive clustering method is the radius of the clusters. During the training phase, the ANFIS model was optimized by varying the radius values to identify the optimum value that results in the highest R and lowest AAPE for the testing dataset. The optimized ANFIS model has a radius of 0.139 and was trained for 200 epochs. The accuracy of the developed ANFIS model in predicting the PEF values is 19.19% AAPE in the training data and 16.04% AAPE in the testing data. Moreover, the correlation coefficient (R) between the predicted and the actual PEF values is 0.715 in the training data and 0.77 in the testing data. **Figure 1** presents the cross plots between the predicted and the actual PEF values. We observe that the ANFIS model underfits the training data because it has higher errors in predicting the PEF values in the training data than in the testing data.

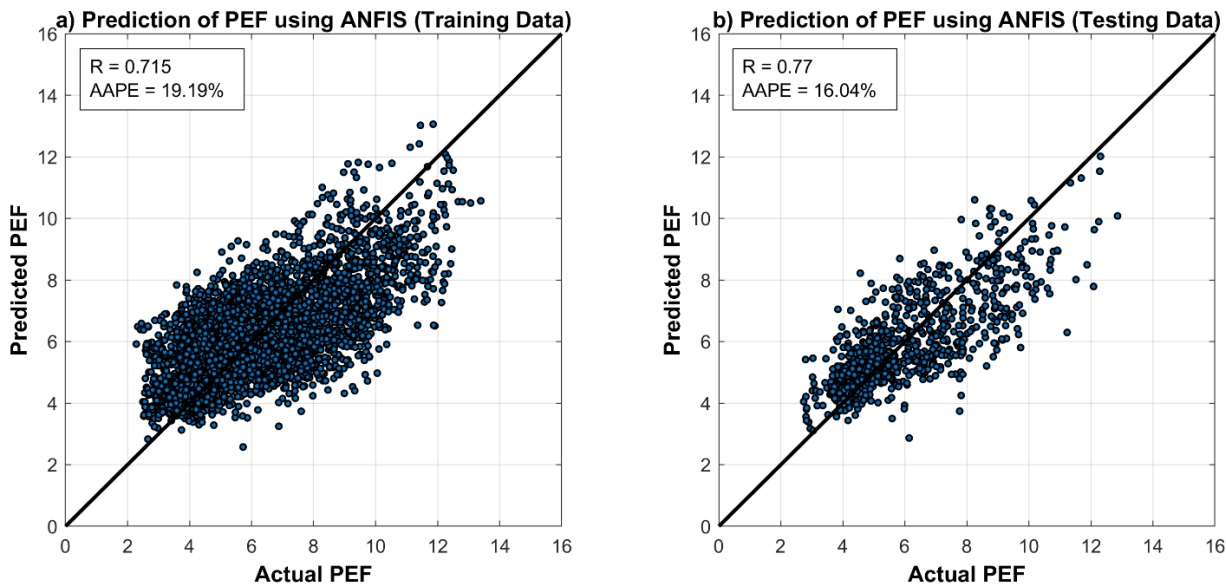
Figure 1: Cross plots between the actual and the predicted PEF values from the ANFIS model.

### 3.2 ANN Results
The accuracy of ANN models is affected by their hyperparameters such as the number of neurons and layers, the transfer (activation) function and the training algorithm that minimizes the loss function by updating the weights of the network. The studied search space of the ANN hyperparameters are the following: the number of hidden layers (1-3), the number of neurons (5-35), the transfer functions (hyperbolic tangent sigmoid, log sigmoid, hard limit, linear, radial basis, and soft max), and the training algorithms (Levenberg–Marquardt, BFGS quasi-Newton, Bayesian, and conjugate gradients).

The developed ANN model was optimized by selecting the optimum combination of those hyperparameters to generate the best accuracy for the testing dataset. The best hyperparameters are 2 hidden layers with 16 and 20 neurons respectively, log sigmoid transfer function, and Bayesian regularization backpropagation. The accuracy of the ANN model in predicting the PEF values is (AAPE=13.32% and R=0.856) in the training data, and (AAPE=14.09% and R=0.815) in the testing data. **Figure 2** shows the cross plots between the actual PEF and the predicted PEF values. Although, the accuracy of the ANN model is better than the ANFIS model, it is still not perfectly accurate.

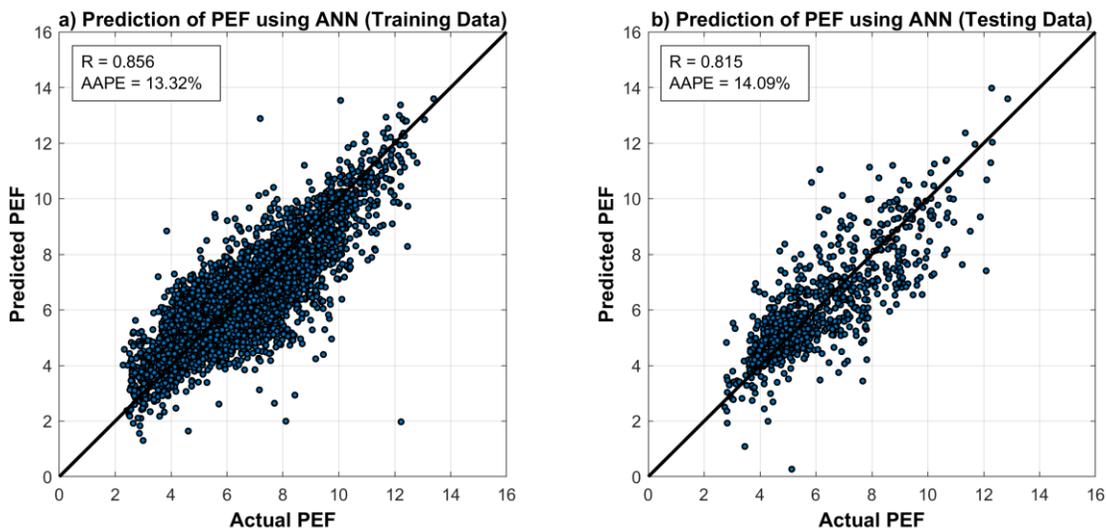
Figure 2: Cross plots between the actual and the predicted PEF values from the ANN model.

## 3.3 GPR Results

Automated machine learning searches for the optimal algorithm (model) for the dataset of interest, by applying different types of methods, while optimizing their hyperparameters for the best results (Feurer et al., 2015). Thus, it increases the efficiency of applying artificial intelligence tools through eliminating the time wasted on manual steps.

The automated learning code compares different models including multi-linear regression, Gaussian kernel regression, decision tree, GPR, ensemble learning (Least-squares gradient boosting and Bootstrap aggregation (random forest)), ANNs and support vector machine (SVM). The code utilizes Bayesian optimization to select the optimal hyperparameters for the models and find the most accurate model for the prediction task (Snoek et al., 2012). The results show that the best model is a GPR model. One additional unique advantage of GPR is that it can incorporate uncertainty, hence accounting for the potential noise in the PEF measurements. This advantage is extremely critical in zones where the PEF values cannot be predicted accurately because it reduces the probability of misinterpreting the predicted PEF log by geoscientists and engineers.

The optimum GPR model has these hyperparameters: sigma=0.032, kernel scale=0.602, constant basis function, and exponential kernel function. It is evident that the GPR model outperforms the ANFIS and the ANN models as summarized in Table 2. The accuracy of the GPR model in predicting the PEF values is (AAPE=0.01% and R=0.999) in the training data, and (AAPE=10.32% and R=0.889) in the testing data. **Figure 3** is the cross plots between the actual PEF and the predicted PEF values. Furthermore, the error of the GPR model in the testing dataset could be further decreased to about (4.6% - 1.9%) using (95% - 99%) confidence interval to model the noise of the dataset.

We observe that the GPR model overfits the training data, which is common to some GPR methods (Mohammed & Cawley, 2017). However, this probably does not reduce its accuracy as can be observed by comparing the GPR model accuracy against the accuracy of the ANN and the ANFIS models using the testing data and the final validation data (next section). In general, weak learners tend to underfit the data (e.g., in our case the ANFIS model), while strong learners tend to overfit the data.

*Table 2 Compares the accuracy of the developed models*

| Method | R (train) | AAPE (train), % | R (test) | AAPE (test), % |
| --- | --- | --- | --- | --- |
| **ANFIS** | 0.715 | 19.19 | 0.77 | 16.04 |
| **ANN** | 0.856 | 13.32 | 0.815 | 14.09 |
| **GPR** | 0.999 | 0.01 | 0.889 | 10.23 |

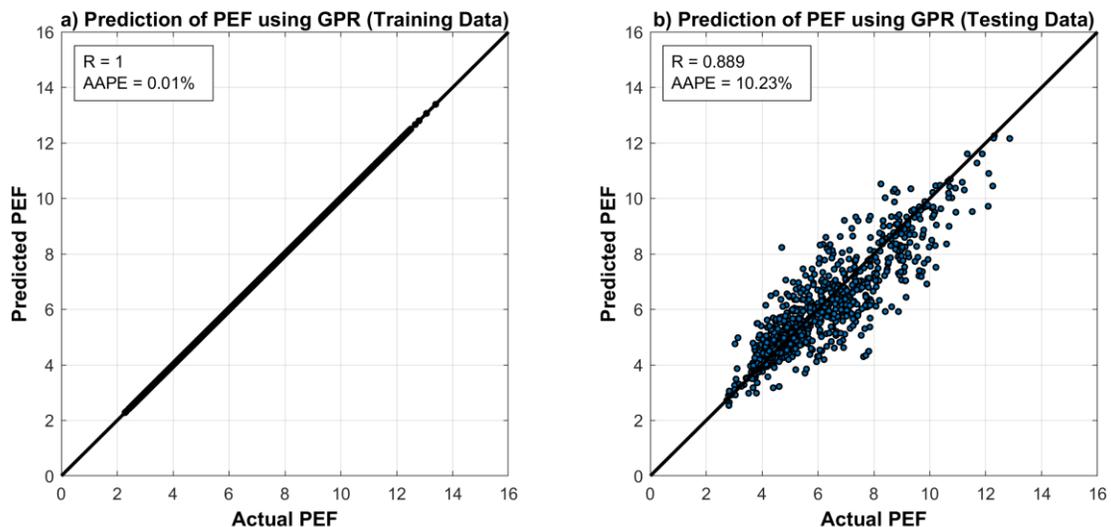

Figure 3: Cross plots between the actual and the predicted PEF values from the GPR model.

## 4 Validation

In this section, the best model (GPR + UQ) is further tested on larger unseen dataset for final validation. This dataset consists of 2603 datapoints. The cross plot between the predicted and the actual PEF values in the validation data is shown in **Figure 4**. The GPR predicts the PEF values with the highest accuracy (AAPE= 9.5% and R=0.894) in the validation data (Table 3), where estimating the uncertainty using 95% confidence (prediction) interval decreases the error to (4.8%). The importance of uncertainty quantification is noticeable in zones where the actual PEF values deviate from the predictions, e.g., depth index from 1080 to 1090 (**Figure 5**). The uncertainty model could also consider the potential noise in the inputs that can be observed from the depth index (1035-1040) in **Figure 5**, where the PEF values are similar but have different confidence estimation because of having different inputs.

Despite that the error of the uncertainty could be further decreased to about 2% using 99% confidence interval, we selected 95% confidence interval as it generates good accuracy with reasonable prediction width. However, in other cases, a different confidence interval (lower or higher) may be selected depending on the knowledge about the predicted parameter, the acceptable range of error and the width of confidence interval.

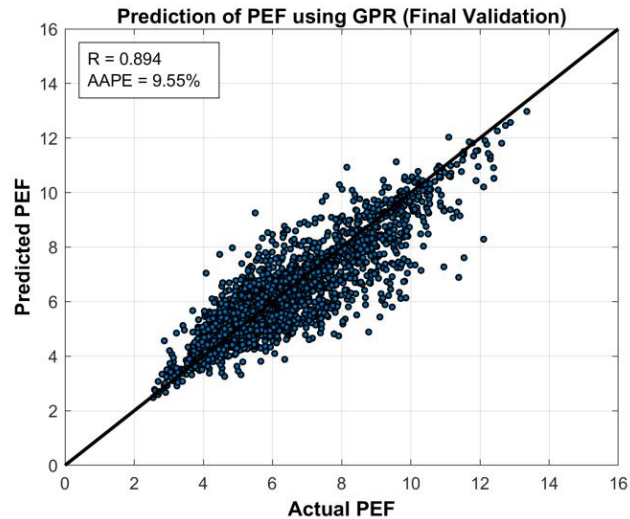

Figure 4: Cross plot between the actual and the predicted PEF values from the GPR model (final validation data).

*Table 3 Compares the accuracy of the developed models using the final validation dataset (2603 datapoints)*

| Method | R (validation) | AAPE (validation), % |
|---|---|---|
| ANFIS | 0.761 | 15.51 |
| ANN | 0.810 | 13.32 |
| GPR | 0.894 | 9.55 |

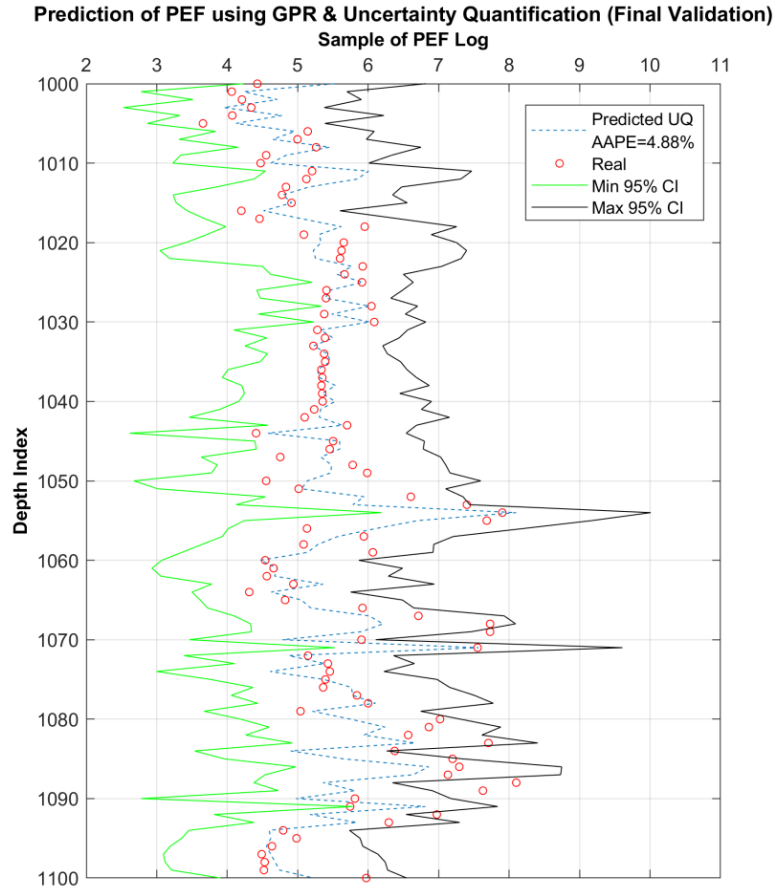

Figure 5: Depth VS the predicted PEF values from the GPR+UQ model (final validation data).

## 5 Conclusion

In this work, we developed various models to predict PEF log. The results show that predicting PEF logs can be challenging because they cannot be predicted accurately for some zones by the common machine learning methods. Similar results have been reported by previous studies. To improve the accuracy of the models, we attempted several data preprocessing and transformation methods, but they were not effective at reducing the prediction error.

Therefore, automated machine learning and uncertainty quantification were proposed for robust prediction of the PEF values in our dataset. The automated learning code recommended the GPR model which outperformed the other machine learning methods in predicting the PEF values. Moreover, uncertainty quantification has allowed further reduction of error by considering the noise in the data, which is crucial for the troublesome reservoir layers. It could be used for risk assessment, hence decreasing the possibilities of misinterpreting the predicted PEF values.

The GPR model generates the most accurate PEF predictions for the dataset under investigation, but could this finding be generalized to other well logs, drilling and geo-mechanical parameters? More investigations may be required to confirm the reliability of the GPR model in predicting PEF logs for different wells, fields, data ranges, and geological layers. Additionally, the causes of the high error in the predicted PEF values using machine learning methods could be studied. This error may be due to the heterogeneity of the dataset, some special reservoir rocks, well (borehole) condition, heavy elements, and noise from the well logging tools (PEF tool).

# References

Akinnikawe, O., Lyne, S., & Roberts, J. (2018). Synthetic Well Log Generation Using Machine Learning Techniques. *Proceedings of the 6th Unconventional Resources Technology Conference*. https://doi.org/10.15530/urtec-2018-2877021

Amir, S. M., Khan, M. R., Panacharoensawad, E., & Kryvenko, S. (2020, October 26). Integration of Petrophysical Log Data with Computational Intelligence for the Development of a Lithology Predictor. *Day 2 Tue, October 27, 2020*. https://doi.org/10.2118/202047-MS

Asante-Okyere, S., Shen, C., Ziggah, Y. Y., Rulegeya, M. M., & Zhu, X. (2018). Investigating the predictive performance of Gaussian process regression in evaluating reservoir porosity and permeability. *Energies*, *11*(12). https://doi.org/10.3390/en11123261

Asoodeh, M., & Shadizadeh, S. R. (2015). The Prediction of Photoelectric Factor, Formation True Resistivity, and Formation Water Saturation from Petrophysical Well Log Data: A Committee Neural Network Approach. *Energy Sources, Part A: Recovery, Utilization, and Environmental Effects*, *37*(5), 557–566. https://doi.org/10.1080/15567036.2011.594859

Atlas, D. (1982). *Well logging and interpretation techniques: The course for home study*. Dresser Atlas.

Bahrpeyma, F., Golchin, B., & Cranganu, C. (2013). Fast fuzzy modeling method to estimate missing logsin hydrocarbon reservoirs. *Journal of Petroleum Science and Engineering*, *112*, 310–321. https://doi.org/10.1016/j.petrol.2013.11.019

Bassiouni, Z., & others. (1994). *Theory, measurement, and interpretation of well logs* (Vol. 4). Henry L. Doherty Memorial Fund of AIME, Society of Petroleum Engineers~….

Blanes de Oliveira, L. A., & de Carvalho Carneiro, C. (2021). Synthetic geochemical well logs generation using ensemble machine learning techniques for the Brazilian pre-salt reservoirs. *Journal of Petroleum Science and Engineering*, *196*, 108080. https://doi.org/10.1016/j.petrol.2020.108080

Chiu, S. L. (1994). Fuzzy Model Identification Based on Cluster Estimation. *Journal of Intelligent and Fuzzy Systems*, *2*(3), 267–278. https://doi.org/10.3233/IFS-1994-2306

Desouky, M., Tariq, Z., Aljawad, M. S., Alhoori, H., Mahmoud, M., & Abdulraheem, A. (2021). Machine Learning-Based Propped Fracture Conductivity Correlations of Several Shale Formations. *ACS Omega*, *6*(29), 18782–18792. https://doi.org/10.1021/acsomega.1c01919

Ellis, D. v, & Singer, J. M. (2007). *Well logging for earth scientists* (Vol. 692). Springer.

Feurer, M., Klein, A., Eggensperger, K., Springenberg, J., Blum, M., & Hutter, F. (2015). Efficient and Robust Automated Machine Learning. In C. Cortes, N. Lawrence, D. Lee, M. Sugiyama, & R. Garnett (Eds.), *Advances in Neural Information Processing Systems* (Vol. 28). Curran Associates, Inc. https://proceedings.neurips.cc/paper/2015/file/11d0e6287202fced83f79975ec59a3a6-Paper.pdf

Gamal, H., Abdelaal, A., & Elkatatny, S. (2021). Machine Learning Models for Equivalent Circulating Density Prediction from Drilling Data. *ACS Omega*, *6*(41), 27430–27442. https://doi.org/10.1021/acsomega.1c04363

Gowida, A., Elkatatny, S., & Gamal, H. (2021). Unconfined compressive strength (UCS) prediction in real-time while drilling using artificial intelligence tools. *Neural Computing and Applications*, *33*(13), 8043–8054. https://doi.org/10.1007/s00521-020-05546-7

Gurney, K. (Kevin N. ). (1997). *An introduction to neural networks*. UCL Press.

Hadavimoghaddam, F., Ostadhassan, M., Sadri, M. A., Bondarenko, T., Chebyshev, I., & Semnani, A. (2021). Prediction of Water Saturation from Well Log Data by Machine Learning Algorithms: Boosting and Super Learner. *Journal of Marine Science and Engineering*, *9*(6), 666. https://doi.org/10.3390/jmse9060666

Hiba, M., Ibrahim, A. F., Elkatatny, S., & Ali, A. (2022). Application of Machine Learning to Predict the Failure Parameters from Conventional Well Logs. *Arabian Journal for Science and Engineering*. https://doi.org/10.1007/s13369-021-06461-2

Hossain, T. M., Watada, J., Aziz, I. A., & Hermana, M. (2020). Machine Learning in Electrofacies

Classification and Subsurface Lithology Interpretation: A Rough Set Theory Approach. *Applied Sciences*, *10*(17), 5940. https://doi.org/10.3390/app10175940

Jang, J.-S. R. (1993). ANFIS: adaptive-network-based fuzzy inference system. *IEEE Transactions on Systems, Man, and Cybernetics*, *23*(3), 665–685. https://doi.org/10.1109/21.256541

Merembayev, T., Kurmangaliyev, D., Bekbauov, B., & Amanbek, Y. (2021). A Comparison of Machine Learning Algorithms in Predicting Lithofacies: Case Studies from Norway and Kazakhstan. *Energies*, *14*(7), 1896. https://doi.org/10.3390/en14071896

Miah, M. I., Zendehboudi, S., & Ahmed, S. (2020). Log data-driven model and feature ranking for water saturation prediction using machine learning approach. *Journal of Petroleum Science and Engineering*, *194*, 107291. https://doi.org/10.1016/j.petrol.2020.107291

Mohammed, R., & Cawley, G. (2017). *Over-Fitting in Model Selection with Gaussian Process Regression* (pp. 192–205). https://doi.org/10.1007/978-3-319-62416-7_14

Onalo, D., Adedigba, S., Oloruntobi, O., Khan, F., James, L. A., & Butt, S. (2020). Data-driven model for shear wave transit time prediction for formation evaluation. *Journal of Petroleum Exploration and Production Technology*, *10*(4), 1429–1447. https://doi.org/10.1007/s13202-020-00843-2

Rasmussen, C., & Williams, C. (2006). *Gaussian processes for machine learning.,(MIT press: Cambridge, MA)*.

Rostamian, A., Heidaryan, E., & Ostadhassan, M. (2022). Evaluation of different machine learning frameworks to predict CNL-FDC-PEF logs via hyperparameters optimization and feature selection. *Journal of Petroleum Science and Engineering*, *208*, 109463. https://doi.org/10.1016/j.petrol.2021.109463

Snoek, J., Larochelle, H., & Adams, R. P. (2012). Practical bayesian optimization of machine learning algorithms. *Advances in Neural Information Processing Systems*, *25*.

Tatsipie, N. R. K., & Sheng, J. J. (2021). Generating pseudo well logs for a part of the upper Bakken using recurrent neural networks. *Journal of Petroleum Science and Engineering*, *200*, 108253. https://doi.org/10.1016/j.petrol.2020.108253

Wu, L., Dong, Z., Li, W., Jing, C., & Qu, B. (2021). Well-logging prediction based on hybrid neural network model. *Energies*, *14*(24). https://doi.org/10.3390/en14248583

Yager, R. R., & Filev, D. P. (1994). Generation of fuzzy rules by mountain clustering. *Journal of Intelligent & Fuzzy Systems*, *2*(3), 209–219.

Yu, Y., Xu, C., Misra, S., Li, W., Ashby, M., Pan, W., Deng, T., Jo, H., Santos, J. E., Fu, L., Wang, C., Kalbekov, A., Suarez, V., Kusumah, E. P., Aviandito, M., Pamadya, Y., & Izadi, H. (2021). Synthetic Sonic Log Generation With Machine Learning: A Contest Summary From Five Methods. *Petrophysics – The SPWLA Journal of Formation Evaluation and Reservoir Description*, *62*(4), 393–406. https://doi.org/10.30632/PJV62N4-2021a4